\documentclass{article}

\PassOptionsToPackage{numbers, compress}{natbib}

\usepackage[preprint]{neurips_2026}

\usepackage[utf8]{inputenc}
\usepackage[T1]{fontenc}
\usepackage{hyperref}
\usepackage{url}
\usepackage{booktabs}
\usepackage{amsfonts}
\usepackage{amsmath}
\usepackage{amssymb}
\usepackage{nicefrac}
\usepackage{microtype}
\usepackage{xcolor}
\usepackage{colortbl}
\usepackage{graphicx}
\usepackage{subcaption}
\usepackage{algorithm}
\usepackage{algorithmic}
\usepackage{multirow}
\usepackage{wrapfig}
\usepackage{enumitem}

\definecolor{codreamblue}{RGB}{31, 119, 180}
\definecolor{evogreen}{RGB}{44, 160, 44}
\definecolor{baselinegray}{RGB}{127, 127, 127}

\newcommand{\evopool}{\textsc{EvoChamber}}
\newcommand{\codream}{\textsc{CoDream}}

\newcommand{\cmark}{\ensuremath{\checkmark}}
\newcommand{\xmark}{\ensuremath{\times}}

\title{\evopool{}: Test-Time Co-evolution of Multi-Agent System at Individual, Team, and Population Scales}

\author{%
  \vspace{-25pt}\\
  \textbf{Yaolun Zhang$^{1,5,*}$,\quad Tianyi Xu$^{2,*}$,\quad Shengyu Dai$^{3}$} \\
  \textbf{Zhenwen Shao$^{3}$,\quad Qingyun Wu$^{4,5}$,\quad Huazheng Wang$^{1,5}$}\vspace{3pt} \\
  $^{1}$Oregon State University \quad\quad $^{2}$University of Wisconsin--Madison \\
  $^{3}$Johnson \& Johnson \quad\quad $^{4}$Pennsylvania State University \quad\quad $^{5}$AG2AI, Inc.\vspace{3pt} \\
  \texttt{\small \{zhanyaol, huazheng.wang\}@oregonstate.edu,\quad txu223@wisc.edu} \\
  \texttt{\small \{SDai9, ZShao5\}@its.jnj.com,\quad qingyun.wu@psu.edu}\vspace{3pt} \\
  $^{*}$Equal contribution.
}

\begin{document}

\maketitle

\begin{abstract}
We argue that multi-agent test-time evolution is not single-agent evolution replicated $N$ times. A single-agent learner can only evolve its own context and memory. A multi-agent system additionally evolves who collaborates, how they collaborate, and how knowledge flows across the population. These components have no single-agent counterpart and can produce phenomena such as emergent specialization. Yet prior test-time methods either confine experiences to individual agents, forfeiting cross-agent learning, or broadcast symmetrically to all agents, erasing the specialization that makes collaboration valuable. We present \evopool{}, a training-free framework that instantiates test-time evolution at three levels over a coevolving agent pool. At its core is \codream{} (\textbf{Co}llaborative \textbf{Dream}ing), a post-task protocol triggered on team failure or disagreement, in which agents collaboratively reflect, distill insights, and route them asymmetrically from strong to weak agents on the failed niche, preserving specialization while filling knowledge gaps. Team-level operators assemble niche-conditioned teams and select collaboration structures online. Population-level lifecycle operators fork, merge, prune, and seed agents under performance pressure. On three heterogeneous task streams with Qwen3-8B, \evopool{} reaches 63.9\% on competition math, 75.7\% on code, and 87.1\% on multi-domain reasoning, outperforming the best baseline by 32\% relative on math and confirming asymmetric cross-agent transfer as the primary driver in ablation. Starting from several identically initialized agents, four to five stable niche specialists spontaneously emerge, a structural signature of multi-agent evolution that no single-agent learner can express. See our code at: \href{https://github.com/Mercury7353/EvoChamber}{https://github.com/Mercury7353/EvoChamber}

\end{abstract}

\section{Introduction}
\label{sec:intro}

Large Language Models (LLMs) \citep{openai2024gpt4technicalreport} excel at reasoning \citep{wei2022cot}, coding, and recall. Multi-agent systems (MAS) built on LLMs assign roles and communication patterns across multiple LLM instances \citep{hong2024metagpt, qian2024chatdevcommunicativeagentssoftware, li2023camel, dylan2024, autogen2024}. Deployed over continual task streams, such systems should improve with experience: breakthroughs should inform later tasks, and recurring task types should be routed to the best-suited agents.

However, evolving a multi-agent system is fundamentally different from evolving a single agent $N$ times in parallel. A single-agent learner, such as Reflexion \citep{shinn2023reflexion} or ExpeL \citep{zhao2023expel}, evolves only one agent's context and memory. A multi-agent system, in contrast, maintains a pool of agents and a strictly richer evolvable state. Beyond the individual level, the state includes a \emph{team} component that determines who collaborates, how they collaborate, and how the joint outcome updates per-agent knowledge. It also includes a \emph{population} component that governs knowledge flow between agents and edits pool membership over time, producing phenomena such as emergent specialization that have no counterpart for a single agent. 

Yet existing work does not instantiate this full state space. Methods that evolve individual agents, including EvoMem \citep{evomem} and MemCollab \citep{memcollab2024}, confine experiences to one agent or broadcast them symmetrically to all agents. The former forfeits cross-agent learning and the latter erases specialization, because every agent receives identical memory regardless of individual strengths. A parallel line of work pursues multi-agent co-improvement through RL fine-tuning \citep{comas2025, maporl2025, mae2025} or offline structure search \citep{evomac2024, zhang2025aflow, evoagent2024}, but these methods operate on fixed agent roles within a single domain and freeze the resulting system at deployment. Neither camp addresses the question: \emph{how can a multi-agent system continuously evolve at test time, across heterogeneous task streams, without gradient updates?}
\begin{figure}[t]
\centering
\includegraphics[width=\linewidth]{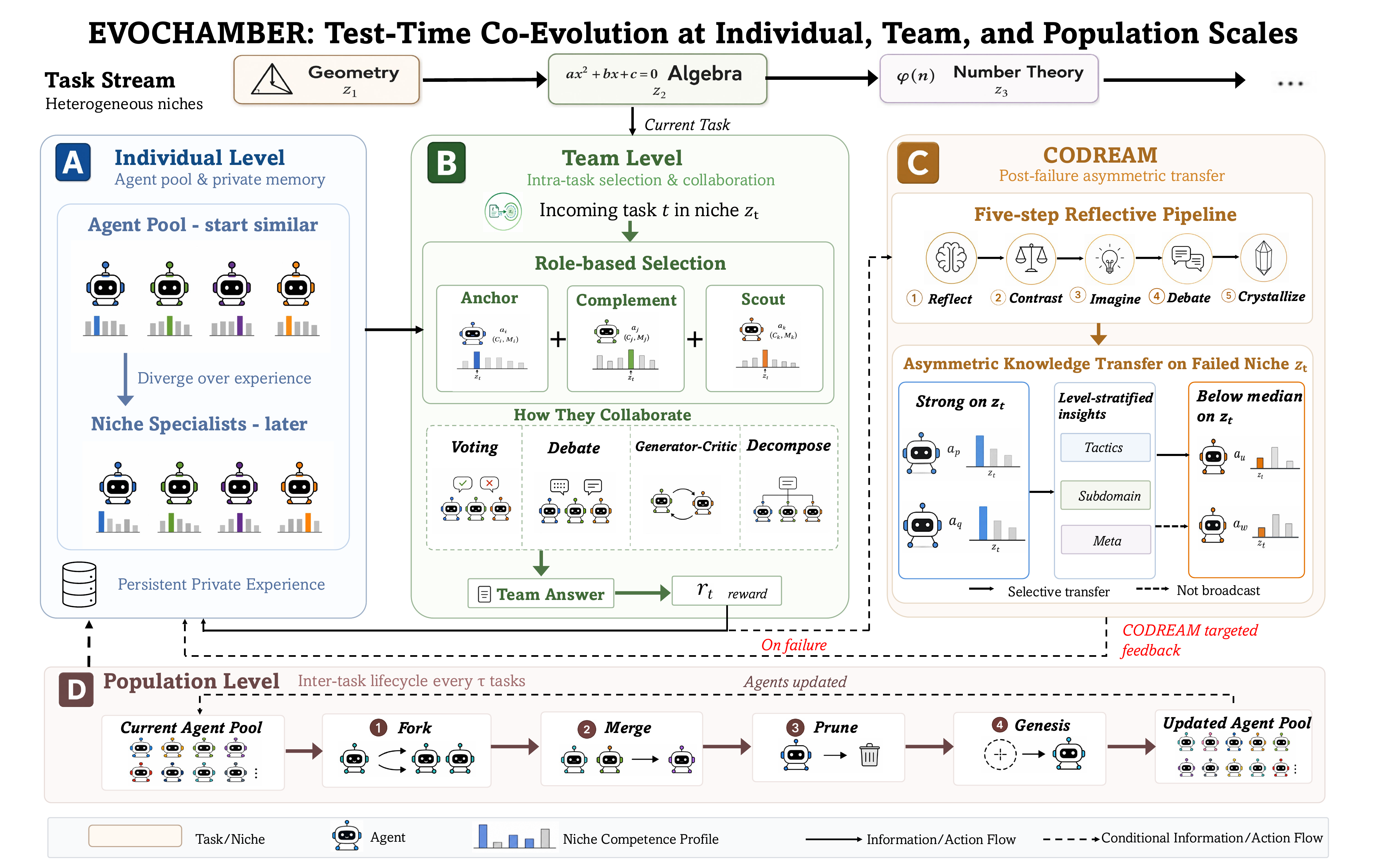}
\vspace{-10pt}
\caption{Overview of \evopool{}. Starting from a pool of $N$ identically initialized agents (\emph{individual level}), a niche-conditioned selector assigns three functional roles, anchor, complement, and scout, and a leader-learned policy selects one of four collaboration structures. The team outcome is attributed as a shared reward (\emph{team level, intra-task}). Between tasks, asymmetric transfer (\codream{}) routes insights from high-fitness to deficit agents, and lifecycle operators fork, merge, prune, and seed new agents to edit pool composition (\emph{population level, inter-task}).}
\vspace{-25pt}
\label{fig:overview}
\end{figure}

To investigate this question, we propose \textbf{\evopool{}}, a training-free framework that instantiates test-time evolution on all three levels over a coevolving agent pool (Fig.~\ref{fig:overview}). At the individual level, every agent accumulates private experience and niche competence estimates. At the team level, a niche-conditioned selector assembles a team of three complementary agents and a leader selects one of four collaboration structures online. At the population level, \codream{} (\textbf{Co}llaborative \textbf{Dream}ing) triggers on team failure or disagreement: agents collaboratively reflect, distill insights, and route them asymmetrically from strong to weak agents on the failed niche, preserving specialization while filling knowledge gaps. Lifecycle operators periodically fork, merge, prune, and seed agents under performance pressure. Table~\ref{tab:related} positions \evopool{} against prior work along the three evolution levels.

\begin{table}[t]
\centering
\small
\renewcommand{\arraystretch}{1.05}
\setlength{\tabcolsep}{3.5pt}
\begin{tabular}{lccccccc}
\toprule
 & \multicolumn{2}{c}{Individual} & \multicolumn{2}{c}{Team (intra-task)} & Population (inter-task) & Online & Training \\
\cmidrule(lr){2-3}\cmidrule(lr){4-5}\cmidrule(lr){6-6}
Method & context & memory & composition & structure & transfer \,/\, pool edit & & free \\
\midrule
Reflexion \citep{shinn2023reflexion}          & \cmark & \cmark & \xmark & \xmark & \xmark \,/\, \xmark & \cmark & \cmark \\
MemCollab \citep{memcollab2024}               & \cmark & \cmark & \xmark & \xmark & sym.\,/\, \xmark & \cmark & \cmark \\
\midrule
CoMAS \citep{comas2025}                       & \cmark$^\ddagger$ & \xmark & \xmark & \xmark & \xmark \,/\, \xmark & \xmark & \xmark \\
MAPoRL \citep{maporl2025}                     & \cmark$^\ddagger$ & \xmark & \xmark & \cmark & \xmark \,/\, \xmark & \xmark & \xmark \\
\midrule
EvoMAC \citep{evomac2024}                     & \cmark & \xmark & \cmark & \cmark & \xmark \,/\, \xmark & \cmark$^\S$ & \cmark \\
AFlow \citep{zhang2025aflow}                       & \xmark & \xmark & \cmark$^\dagger$ & \cmark$^\dagger$ & \xmark \,/\, \xmark & \xmark & \cmark \\
\midrule
\textbf{\evopool{} (ours)}                    & \cmark & \cmark & \cmark & \cmark & \cmark \,/\, \cmark & \cmark & \cmark \\
\bottomrule
\end{tabular}
\vspace{0.3em}

\footnotesize{$\ddagger$ CoMAS and MAPoRL update weights via RL rather than evolving at test time. $\S$ EvoMAC adapts within one task only. $\dagger$ AFlow's structure search is offline and frozen at inference.}
\vspace{-1pt}
\caption{Evolution levels activated by representative methods. \evopool{} is the first to activate all three levels online without training. See \S\ref{sec:related} for extended discussion.}
\vspace{-20pt}
\label{tab:related}
\end{table}

We evaluate \evopool{} on three heterogeneous task streams and two model families. With Qwen3-8B, \evopool{} reaches $63.9\%$ on Hard Math, $75.7\%$ on Hard Code, and $87.1\%$ on AFlow-Stream, outperforming the best baseline MemCollab by $32\%$ relative on math and achieving a $5\times$ improvement on CodeContests over a single agent. Gains are largest in the hardest regimes and transfer to GPT-4.1-mini. Ablations that disable the team or population level yield level-specific drops, with the single largest drop of $-10.8\%$ from removing \codream{}, confirming asymmetric cross-agent transfer as the primary driver. Beyond aggregate accuracy, we observe a signature that is structurally impossible for any single-agent learner: starting from several identically initialized agents, four to five stable niche specialists spontaneously emerge, and this pattern is reproducible across random seeds even though the identity of each specialist changes.

\section{Related Work}
\label{sec:related}

\textbf{Static multi-agent systems.}
AutoGen~\citep{autogen2024}, MetaGPT~\citep{hong2024metagpt}, CAMEL~\citep{li2023camel}, DyLAN~\citep{dylan2024}, AgentVerse~\citep{chen2024agentverse}, and Mixture-of-Agents~\citep{wang2024moa} assign fixed or dynamically grouped roles, but agent knowledge cannot evolve with the task stream.
Multi-agent debate~\citep{du2024debate, liang2024mad} and test-time reasoning enhancements~\citep{yao2023tot, snell2025testtimescaling} improve answer quality but carry no persistent state across tasks.
AFlow~\citep{zhang2025aflow}, Archon~\citep{saadfalcon2025archon}, ADAS~\citep{hu2025adas}, and ScoreFlow~\citep{ma2025scoreflow} discover workflows or agent architectures offline via search, while GPTSwarm~\citep{zhuge2024gptswarm} and MacNet~\citep{qian2025macnet} optimize multi-agent graphs via gradient signals, yet the result is frozen at inference time.
EvoMAC~\citep{evomac2024} adapts agent interactions within a single task but does not carry experience across tasks.
\evopool{} is complementary: where automated design optimizes \emph{workflow graphs} offline, \evopool{} evolves \emph{agent content} online.

\textbf{Individual agent memory.}
Self-Refine~\citep{madaan2023selfrefine} iterates on a single agent's output through self-feedback, Reflexion~\citep{shinn2023reflexion} accumulates self-critiques, ExpeL~\citep{zhao2023expel} extracts reusable insights from trajectories, and AgentNet~\citep{agentnet2024} equips agents with personal RAG stores.
EvoMem~\citep{evomem} extends Reflexion-style memory to a pool setting.
All improve individual agents but provide no mechanism for one agent's learning to transfer to another, which is critical at low success rates where individual memory accumulates mostly failures.

\textbf{Symmetric shared memory.}
MemCollab~\citep{memcollab2024} distills team trajectories into a shared store broadcast to all agents, enabling collective learning, but the sharing is symmetric: every agent receives identical memory regardless of individual strengths, conflating domain-specific strategies and destroying specialization.
\evopool{}'s \codream{} addresses this through asymmetric, gap-targeted distillation that routes insights only to deficit agents.

\textbf{Gradient-based co-evolution.}
CoMAS~\citep{comas2025} co-evolves agents via interaction rewards, MAPoRL~\citep{maporl2025} applies multi-agent post-co-training with RL, MAE~\citep{mae2025} pursues LLM self-improvement through co-evolution, and MAS\textsuperscript{2}~\citep{zhang2026mas2} specializes agents via DPO.
These methods require gradient updates on a static training distribution.
\evopool{} achieves comparable qualitative goals through inference-time prompt evolution alone. No prior work simultaneously achieves pool-level persistent state, verified asymmetric cross-agent distillation, and structural pool evolution, all without gradient updates and all online (Appendix~\ref{app:positioning}).
\section{Method}
\label{sec:method}

\subsection{Problem Formulation and the Solve-Evolve Loop}
\label{sec:problem}

Let $\mathcal{T} = (t_1, \ldots, t_T)$ be an online stream of tasks drawn from $K$ heterogeneous niches, with per-task niche label $z_t$ and reward $r_t \in [0,1]$. The objective is to maximize $\sum_t r_t$ by evolving system state.

\textbf{The per-task loop.} For each task $t$, \evopool{} (i) \emph{selects a team} of three agents with roles anchor, complement, scout. (ii) The anchor (also leader) chooses structure $L_t$ from its experiences. (iii) The team \emph{executes} $L_t$, scoring as $r_t$. (iv) $r_t$ \emph{propagates} as a shared reward, updating per-agent competence and pool-wide pair synergy. On failure or disagreement, a post-hoc \codream{} session emits insights to deficit agents. (v) Every $\tau$ tasks, lifecycle operators (fork, merge, prune, genesis) edit pool membership. 


\subsection{What Evolves: Three-Level State Decomposition}
\label{sec:state}

A single-agent learner evolves only $\theta_t^{\text{SA}} = (C_t, M_t)$, where $C_t$ is the working context and $M_t$ is the persistent store retrieved into $C_t$. A multi-agent system maintains a pool $\mathcal{P}_t = \{a_1, \ldots, a_{|\mathcal{P}_t|}\}$ and a richer evolvable state
\begin{equation}
\theta_t^{\text{MAS}} \;=\; \underbrace{\{(C_t^i, M_t^i)\}_{i \in \mathcal{P}_t}}_{\text{individual}} \;\oplus\; \underbrace{(T_t, L_t)}_{\text{team (intra-task)}} \;\oplus\; \underbrace{(\Sigma_t, \Omega_t, \mathcal{P}_t)}_{\text{population (inter-task)}},
\label{eq:state}
\end{equation}
where $T_t$ is the size-$k$ team selected for task $t$ and $L_t$ is the collaboration structure used to combine its outputs. The remaining three quantities persist across tasks and drive how teams are formed.

\textbf{Pair-wise synergy} $\Sigma_t$ captures whether agents $i$ and $j$ work well together on niche $z$, a question no per-agent statistic can answer. We maintain $\Sigma_t[i,j,z] = \sigma_{ij}(z)$ as the running mean team reward over past niche-$z$ tasks in which $i$ and $j$ co-participated. Composition (\S\ref{sec:team}) reads $\Sigma_t$ to favor complements with high prior synergy with the anchor.

\textbf{Pair-wise style overlap} $\Omega_t$ prevents teams of strong but redundant agents. We define $\Omega_t[i,j] = \omega_{ij} = \cos(\vec{q}_i, \vec{q}_j)$, the cosine similarity between niche-competence vectors $\vec{q}_i = (q_i(z_1), \ldots, q_i(z_K))$. Composition penalizes high $\omega$ when adding members, biasing teams toward complementary skill profiles. $\Omega_t$ is derived from $\{\vec{q}_i\}$ and requires no separate update.

\textbf{Mutable roster} $\mathcal{P}_t$ is the set of active agents, with $|\mathcal{P}_t| \gg k$ so that selection has room to maneuver. $\mathcal{P}_t$ is itself evolvable: lifecycle operators (\S\ref{sec:population}) periodically fork, merge, prune, and seed agents, so the pool's \emph{shape}, not just its members' memories, adapts to the task stream.

\begin{figure}[t]
\centering
\includegraphics[width=\linewidth]{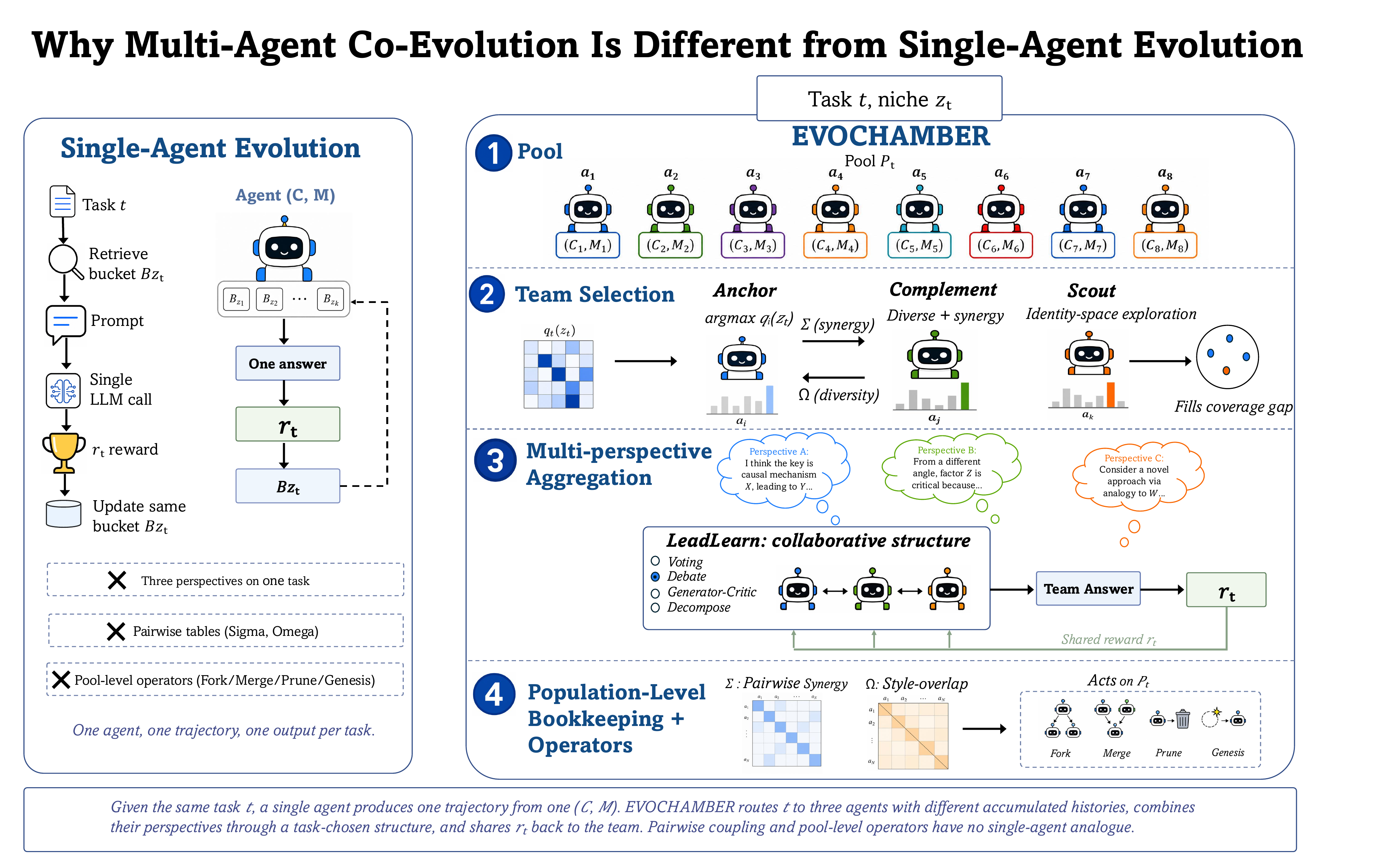}
\vspace{-15pt}
\caption{Same task $t$, two treatments. \textbf{Left:} a single agent produces one trajectory from one memory store. \textbf{Right:} three agents drawn from a pool of $N$ heterogeneous histories, aggregated by a leader-chosen structure. Shared reward updates $(\Sigma, \Omega)$, and lifecycle operators edit $\mathcal{P}_t$ every $\tau$ tasks.}
\vspace{-17pt}
\label{fig:sa-vs-evochamber}
\end{figure}

Figure~\ref{fig:sa-vs-evochamber} illustrates the gap on a single task: a single $(C, M)$ produces one trajectory and one answer, while the multi-agent state routes the same task to three agents with different accumulated histories, aggregates their perspectives through a task-chosen structure, and updates $(\Sigma, \Omega, \mathcal{P})$ as a side effect. The next three subsections detail each level.
\subsection{Individual-Level Evolution}
\label{sec:individual}
The individual level maintains each agent's private knowledge: its
accumulated experience and niche competence. 

\textbf{Experience archive.}
After each task in which $a_i$ participates, $a_i$ reflects on its intermediate outputs, the team's answer, and the reward. The reflection produces two lessons at different granularities: a \emph{subtask-level} lesson indexed by the niche label $z_t$, and a \emph{cross-domain meta-insight} not tied to any niche. Subtask lessons are bucketed by niche, meta-insights form one pool, and both grow with the agent's full history without capacity limit. At solve time, $a_i$ retrieves the top-$k$ entries from its niche-$z_t$ bucket and meta-insight pool by cosine similarity over task embeddings, and prepends them to the prompt. This reflection is independent of LeadLearn (\S\ref{sec:team}): one tracks how to solve, the other how to organize collaboration.

\textbf{Niche competence.} 
Beyond textual experience, each agent also tracks a scalar competence 
$q_i(z) \in [0,1]$ estimating its expected reward on niche-$z$ tasks. 
After each task with outcome $r_t \in [0,1]$, we update via EWMA:
\begin{equation}
    q_i(z) \leftarrow (1-\alpha)\,q_i(z) + \alpha\,r_t,
\end{equation}
initialized at $q_i(z) = 0.5$. EWMA is preferred over a running mean
because competence is non-stationary as the agent's experience and
teammates evolve, so recent outcomes carry more signal.


\subsection{Team-Level Evolution}
\label{sec:team}
The team level assembles an agent team for each incoming task and
decides how they collaborate. Individual heterogeneity emerges here:
agents diverge only because team selection routes them to different task histories.

\textbf{Composition: anchor, complement, scout.}
Picking the top three agents by $q_i(z_t)$ collapses diversity: strong
agents accumulate all experience, weak agents never participate, and the
pool loses the variety that lifecycle operators rely on. We therefore
decompose the team into three roles with distinct selection rules.
The \textbf{anchor} is the niche's current best performer,
\begin{equation}
    a_t = \arg\max_{i \in \mathcal{P}} q_i(z_t),
\end{equation}
with ties broken uniformly at random. It also serves as leader, avoiding a separate election. 
The \textbf{complement} is then drawn from the remaining pool 
$\mathcal{P} \setminus \{a_t\}$ to supply capability the anchor lacks:
\begin{equation}
    c_t = \arg\max_{i \in \mathcal{P} \setminus \{a_t\}} 
    \;\lambda_q\, q_i(z_t) 
    + \lambda_\sigma\, \sigma_{i,a_t}(z_t) 
    + \lambda_\omega\, (1-\omega_{i,a_t}),
\end{equation}
which jointly rewards own competence on $z_t$, prior synergy with the 
anchor on $z_t$, and stylistic distinctness from the anchor. 
The \textbf{scout} is drawn from the rest to enforce exploration and 
diversity:
\begin{equation}
    s_t = \arg\max_{k \in \mathcal{P} \setminus \{a_t, c_t\}} 
    \;\lambda_u\, u_k(z_t) 
    + \lambda_d\, \big(1 - \bar\omega_{k,\{a_t,c_t\}}\big),
\end{equation}
where $u_k(z_t) = 1/(1+n_k(z_t))$ favors agents under-exposed on niche 
$z_t$ and $\bar\omega_{k,\{a_t,c_t\}}$ is the mean style overlap with 
the two already-selected agents. This prevents collapse onto a few dominant members by ensuring every agent periodically receives task experience.
All weight coefficients $\lambda_{(\cdot)}$ are fixed across experiments.

\textbf{Structure: LeadLearn.} 
\label{sec:leadlearn}
Once the team is fixed, the leader chooses a collaboration structure 
$L_t$ from \{\textbf{voting}, \textbf{debate}, \textbf{generator-critic}, 
\textbf{decompose}\}. No single structure dominates across niches, so the leader learns
this choice online. The pool maintains a \emph{shared} experience bank
of past leadership rounds, each entry a tuple
\emph{(team profile, task profile, structure, outcome, reflection)}.
Sharing the bank lets (team, task)$\to$structure meta-knowledge accumulate as the anchor rotates. At decision time, the leader forms a query vector 
$\xi_t$ from the niche label and team competence profile, retrieves 
top-$k$ entries by cosine similarity, and conditions the backbone LLM 
on these to propose $L_t$. After the task, the leader appends a new tuple with a short natural-language note on why $L_t$ succeeded or failed, giving the bank a richer signal than scalar rewards alone.

\textbf{Updates.} After each task, all three agents update $q_i(z_t)$ via EWMA and increment $n_i(z_t)$. Pair synergy is updated analogously,
\begin{equation}
    \sigma_{ij}(z_t) \leftarrow (1-\beta)\,\sigma_{ij}(z_t) + \beta\, r_t,
\end{equation}
since pair compatibility is non-stationary as the agents evolve.
The style overlap $\omega_{ij}$ is recomputed from the updated skill profiles.
The leader's LeadLearn update is described above.

\subsection{Population-Level Evolution}
\label{sec:population}

Two gaps remain after the individual and team levels: a useful lesson discovered by a strong agent stays inside that agent, and the pool's roster is itself a state that should evolve as new task types appear or old strengths become redundant. \codream{} addresses the first by routing knowledge between existing agents, while the lifecycle edits pool membership.

\textbf{\codream{}: knowledge flow without dilution.} A session fires whenever the team fails, either because the mean reward falls below threshold $\theta$ or because members disagree. The three team members run a five-phase reasoning loop: \emph{Reflect} lets each member privately diagnose what went right or wrong in its own attempt. \emph{Contrast} pairs failing members with successful ones to extract a delta, what the successful approach did differently. \emph{Imagine} turns those deltas into hypothetical strategies tagged with the niches they might apply to. \emph{Debate} has the members cross-critique each other's proposals, dropping weak ones. \emph{Crystallize} converts surviving proposals into structured insights, each tagged with a level (task-local, subdomain-scoped, or cross-domain) and a niche scope. The insight is then written into every agent whose competence on that niche falls below the pool median. Strong agents thus produce knowledge while weak ones consume it, sharpening specialization rather than diluting it, the failure mode of symmetric broadcast~\citep{memcollab2024}.

\textbf{Lifecycle: the pool roster as a variable.} Every $\tau$ tasks the system inspects the pool and applies four operators, each targeting a different pathology of a static roster. \textbf{Genesis} fills coverage gaps: when a recurring task type has no specialist, a fresh agent is spawned from the most generalist parent with a persona aimed at the new type. \textbf{Fork} provides specialist headroom: when an agent dominates one task type, the system clones it with a persona mutation that further emphasizes that subdomain, preserving the parent. \textbf{Merge} removes duplication: when two agents have nearly identical skill profiles, they are consolidated, freeing a slot. \textbf{Prune} removes dead weight: an agent whose recent score lags the pool mean over a sustained window is retired. A fifth operator, \emph{specialize}, nudges a high-performing agent's persona toward its dominant niche without changing the roster, so future selections sharpen the same agent rather than scattering experience.

The two halves of population-level evolution are decoupled: \codream{} continuously moves \emph{what is known} between agents, while the lifecycle periodically reshapes \emph{which agents exist}. Because $|\mathcal{P}| > k$, unused agents retain their state, so the pool carries old specialists alongside newly seeded ones without overwriting either.
\section{Experiments}
\label{sec:experiments}

We evaluate \evopool{} on three heterogeneous task streams and two model families, then verify robustness, decompose contributions via ablations, and analyze how the pool evolves.

\subsection{Setup}
\label{sec:setup}

\textbf{Datasets.} We construct three task streams that span different difficulty regimes and domain compositions. The Hard Math Stream combines 262 MATH~\citep{hendrycks2021math} Level~4/5 problems with 30 problems from each of AIME 2022--2025, totaling 382 tasks. The Hard Code Stream contains 257 MBPP+~\citep{austin2021mbpp, liu2024evalplus}, and 165 CodeContests~\citep{li2022alphacode} problems, totaling 422 tasks that test whether debugging experience transfers across problem classes. The AFlow-Stream presents six domains in sequential 100-task blocks: GSM8K~\citep{cobbe2021gsm8k} $\rightarrow$ HotpotQA~\citep{yang2018hotpotqa} $\rightarrow$ MBPP $\rightarrow$ MATH $\rightarrow$ HumanEval~\citep{chen2021codex} $\rightarrow$ DROP~\citep{dua2019drop}, totaling 600 tasks that test adaptation under cross-block domain shifts. Each task carries a niche label $z_t$ derived from its dataset metadata: MATH Level~4/5 vs.\ each AIME year for Hard Math, source benchmark for Hard Code, and domain block for AFlow-Stream. These labels index the per-niche competence statistics in \S\ref{sec:method}.

\textbf{Baselines.} We compare against methods spanning different evolution levels. As no-evolution references, we include a stateless single agent (SA) and majority voting (SC, $k{=}5$)~\citep{wang2023selfconsistency} as a compute-matched comparison. EvoMem~\citep{evomem} and AgentNet~\citep{agentnet2024} evolve per-agent memory without cross-agent transfer, while MemCollab~\citep{memcollab2024} extends this with symmetric pairwise sharing. DyLAN~\citep{dylan2024} adapts collaboration structures at inference time but maintains no cross-task state. All multi-agent baselines use $k{=}3$ agents to match our team size.

\textbf{Implementation.} \evopool{} uses $N{=}20$ identically initialized agents with team size $k{=}3$. The primary backbone is Qwen3-8B~\citep{qwen3} served by 1 H100 GPU, and GPT-4.1-mini~\citep{openai2025gpt41} from API for cross-backbone validation. A single hyperparameter configuration is used across all three streams and both model families with no per-benchmark tuning. See Appendix~\ref{app:hyperparams}.

\textbf{Metrics.} We report accuracy per stream: exact match for math, pass@1 for code, and F1 for QA.

\subsection{Main Results}
\label{sec:main-results}

\begin{table}[t]
\centering
\caption{Hard Math Stream accuracy on Qwen3-8B. math\_hard: 262 MATH Level~4/5; AIME'22--'25: 30 problems each; Overall: micro-average over 382 tasks.}
\label{tab:math}
\setlength{\tabcolsep}{4.5pt}
\tiny
\begin{tabular}{lcccccc}
\toprule
Method & math\_hard & AIME'22 & AIME'23 & AIME'24 & AIME'25 & Overall \\
\midrule
SA              & 0.374 & 0.133 & 0.100 & 0.133 & 0.167 & 0.298 \\
SC ($k{=}5$)    & 0.542 & 0.033 & 0.133 & 0.233 & 0.067 & 0.390 \\
DyLAN           & 0.542 & 0.033 & 0.067 & 0.167 & 0.133 & 0.403 \\
AgentNet        & 0.496 & 0.267 & 0.167 & 0.200 & 0.267 & 0.414 \\
EvoMem          & 0.553 & 0.133 & 0.133 & 0.267 & \textbf{0.300} & 0.445 \\
MemCollab       & 0.603 & 0.233 & 0.167 & 0.267 & 0.233 & 0.484 \\
\midrule
\rowcolor{cyan!5} \textbf{\evopool{}} & \textbf{0.763} & \textbf{0.400} & \textbf{0.333} & \textbf{0.433} & \textbf{0.300} & \textbf{0.639} \\
\bottomrule
\end{tabular}
\vspace{-18pt}
\end{table}

Tables~\ref{tab:math}--\ref{tab:aflow} tell a consistent story across three streams: \evopool{} improves most where single-agent methods struggle, the advantage grows with task difficulty, and cross-agent knowledge transfer is what closes the gap.

\textbf{Largest gains on the hardest tasks.} On the Hard Math Stream (Table~\ref{tab:math}), \evopool{} reaches 0.639 overall, outperforming MemCollab by 32\% relative and doubling the single-agent baseline. The gain concentrates where it matters most: $+$0.160 on math\_hard and $+$0.167 on AIME'24. SC collapses on AIME to 0.067 because majority voting overrides rare correct outputs when per-agent accuracy is below 50\%. \evopool{} avoids this by routing through a niche-competent anchor under a leader-selected structure.

\begin{table}[t]
\centering
\caption{Accuracy on Hard Code Stream and AFlow-Stream. The HumanEval column is omitted from Hard Code as all methods score 1.000; Overall is the micro-average over all 586 tasks including HumanEval. The full breakdown is in Appendix~\ref{app:code}.}
\label{tab:code}\label{tab:aflow}
\tiny

\begin{minipage}[t]{0.34\linewidth}
\centering
{\scriptsize\textbf{(a) Hard Code Stream}}\\[2pt]
\setlength{\tabcolsep}{3pt}
\begin{tabular}{lccc}
\toprule
Method & MBPP+ & CC & Overall \\
\midrule
SA                    & 0.842 & 0.068 & 0.667 \\
SC ($k{=}5$)          & 0.849 & 0.198 & 0.708 \\
DyLAN                 & 0.825 & 0.189 & 0.695 \\
AgentNet              & \textbf{0.887} & 0.102 & 0.698 \\
EvoMem                & 0.885 & 0.027 & 0.672 \\
MemCollab             & 0.870 & 0.084 & 0.682 \\
\midrule
\rowcolor{cyan!5} \textbf{\evopool{}} & 0.861 & \textbf{0.352} & \textbf{0.757} \\
\bottomrule
\end{tabular}
\end{minipage}%
\hfill
\begin{minipage}[t]{0.64\linewidth}
\centering
{\scriptsize\textbf{(b) AFlow-Stream}}\\[2pt]
\setlength{\tabcolsep}{3pt}
\begin{tabular}{lccccccc}
\toprule
Method & GSM8K & HotpotQA & MBPP & MATH & HE & DROP & Overall \\
\midrule
SA                & 0.960 & 0.791 & 0.780 & 0.780 & 0.800 & 0.800 & 0.819 \\
SC ($k{=}5$)      & 0.890 & 0.778 & 0.560 & 0.610 & 0.410 & 0.690 & 0.656 \\
DyLAN             & 0.670 & 0.888 & 0.690 & 0.620 & 0.830 & 0.840 & 0.756 \\
AgentNet          & 0.970 & 0.820 & 0.793 & 0.680 & \textbf{0.900} & 0.800 & 0.827 \\
EvoMem            & 0.940 & 0.892 & 0.817 & 0.660 & 0.880 & 0.850 & 0.840 \\
MemCollab         & 0.960 & 0.847 & 0.793 & 0.660 & 0.890 & 0.840 & 0.832 \\
\midrule
\rowcolor{cyan!5} \textbf{\evopool{}} & \textbf{0.980} & \textbf{0.895} & \textbf{0.843} & \textbf{0.820} & 0.830 & \textbf{0.860} & \textbf{0.871} \\
\bottomrule
\end{tabular}
\end{minipage}
\vspace{-18pt}
\end{table}

\textbf{Experience transfers across difficulty levels.} On the Hard Code Stream (Table~\ref{tab:code}), MBPP+ saturates near 0.85 for all multi-agent methods. The discriminative subset is CodeContests, where \evopool{} reaches 0.352, a $5\times$ improvement over a single agent. Debugging patterns learned on easier MBPP+ problems accumulate in agent profiles and propagate to deficit agents via \codream{}, carrying over to the harder CodeContests problems. EvoMem and MemCollab score below SA on CodeContests at 0.027 and 0.084 respectively, suggesting that individual-level or symmetric memory alone introduces noise that hurts on the hardest problems without the niche-conditioned routing that \codream{} provides.

\textbf{Cross-domain adaptation across sequential domain blocks.} On AFlow-Stream (Table~\ref{tab:aflow}), where six domains arrive in sequential 100-task blocks, \evopool{} reaches 0.871, ahead of EvoMem at 0.840 and MemCollab at 0.832. \evopool{} wins or ties on five of six domains, with the largest gains on MATH and MBPP where cross-agent coordination matters most. This stream tests exactly the scenario our three-level evolution is designed for: agents must specialize within domains while transferring metacognitive strategies across them.

\begin{table}[t]
\centering
\caption{Cross-backbone validation on the Hard Math Stream (top) and AFlow-Stream (bottom) under GPT-4.1-mini. The same hyperparameter configuration is used across both backbones and both streams.}
\label{tab:backbone}
\tiny
\setlength{\tabcolsep}{4pt}
\begin{tabular}{llcccccccc}
\toprule
Backbone (Stream) & Method & \multicolumn{5}{c}{Subset Accuracy} & Overall & $\Delta$ vs SA \\
\cmidrule(lr){3-7}
& & math\_hard & AIME'22 & AIME'23 & AIME'24 & AIME'25 & & \\
\midrule
\multirow{4}{*}{GPT-4.1-mini (Hard Math)}
  & SA                   & 0.824 & 0.400 & 0.300 & 0.333 & 0.367 & 0.675 & --- \\
  & MemCollab            & 0.878 & 0.533 & 0.433 & 0.567 & 0.533 & 0.764 & $+$0.075 \\
  & EvoMem               & 0.882 & 0.600 & 0.367 & 0.500 & 0.467 & 0.757 & $+$0.068 \\
  \rowcolor{cyan!5}
  & \textbf{\evopool{}}  & \textbf{0.889} & \textbf{0.600} & \textbf{0.567} & 0.533 & \textbf{0.567} & \textbf{0.796} & $+$\textbf{0.107} \\
\midrule
& & GSM8K & HotpotQA & MBPP & MATH & HE & DROP & Overall & $\Delta$ vs SA \\
\midrule
\multirow{4}{*}{GPT-4.1-mini (AFlow-Stream)}
  & SA                   & 0.940 & 0.847 & 0.887 & 0.800 & 0.940 & 0.800 & 0.869 & --- \\
  & MemCollab            & 0.950 & 0.864 & 0.910 & 0.680 & 0.940 & 0.850 & 0.866 & $-$0.003 \\
  & EvoMem               & 0.940 & 0.896 & 0.910 & 0.680 & 0.950 & 0.860 & 0.873 & $+$0.004 \\
  \rowcolor{cyan!5}
  & \textbf{\evopool{}}  & \textbf{0.950} & 0.878 & \textbf{0.960} & \textbf{0.820} & 0.940 & 0.780 & \textbf{0.888} & $+$\textbf{0.019} \\
\bottomrule
\end{tabular}
\vspace{-15pt}
\end{table}

\textbf{Gains transfer across backbones and streams.} Table~\ref{tab:backbone} shows that the same hyperparameter configuration lifts \evopool{} above all baselines on both backbones and both streams. The relative lift is larger when the backbone is weaker or the regime is harder: $+$0.341 on Qwen3-8B Hard Math, $+$0.107 on GPT-4.1-mini Hard Math, and $+$0.019 on GPT-4.1-mini AFlow, because GPT-4.1-mini's SA baseline on AFlow already reaches 0.869, leaving little headroom. \evopool{} remains the best method on both GPT-4.1-mini streams.


\subsection{Ablation Studies}
\label{sec:ablation}

\begin{table}[t]
\centering
\caption{(a) Ablation on AFlow-Stream: each row disables one of the method-level innovations \evopool{} introduces, mapped to the Method subsection that describes it. (b) Hard Math Stream under two random task permutations.}
\label{tab:ablation}\label{tab:robustness}
\tiny

\begin{minipage}[t]{0.62\linewidth}
\centering
{\scriptsize\textbf{(a) Ablation on AFlow-Stream}}\\[2pt]
\setlength{\tabcolsep}{3pt}
\begin{tabular}{llcc}
\toprule
Innovation (\S) & Configuration & Acc. & $\Delta$ \\
\midrule
\rowcolor{cyan!5} --- & \evopool{} (full)                                  & 0.871 & --- \\
\midrule
Team composition (\S\ref{sec:team})  & Random team (no niche-conditioned selector)   & 0.847 & $-$0.024 \\
Team structure (\S\ref{sec:team})    & LeadLearn disabled (forced voting)             & 0.841 & $-$0.030 \\
Cross-agent transfer (\S\ref{sec:population}) & $-$ \codream{} entirely                  & 0.763 & $-$0.108 \\
\bottomrule
\end{tabular}
\end{minipage}%
\hfill
\begin{minipage}[t]{0.36\linewidth}
\centering
{\scriptsize\textbf{(b) Hard Math Stream (permutations)}}\\[2pt]
\setlength{\tabcolsep}{3pt}
\begin{tabular}{lccc}
\toprule
Condition & SA & \evopool{} & $\Delta$ \\
\midrule
Default (fixed order) & 0.298 & 0.639 & $+$0.341 \\
Shuffle (seed 42)     & 0.298 & 0.655 & $+$0.357 \\
Shuffle (seed 123)    & 0.298 & 0.662 & $+$0.364 \\
\bottomrule
\end{tabular}
\end{minipage}
\vspace{-10pt}
\end{table}

Table~\ref{tab:ablation} decomposes contributions by evolution level on AFlow-Stream. The single largest drop comes from removing \codream{} entirely: $-$0.108, establishing asymmetric cross-agent transfer as the primary driver of collective learning. The effect is sharpest on dependent-reasoning domains where cross-agent coordination is essential, with HotpotQA dropping from 0.895 to 0.572 and DROP from 0.860 to 0.480.  At the team level, disabling the niche-conditioned selector and disabling LeadLearn each produce independent drops of $-$0.024 and $-$0.030 respectively, confirming that team composition and team structure contribute separately. All innovations are non-redundant, and the gains decompose cleanly across the three evolution levels.

We also analyze the robustness of \evopool{}. Table~\ref{tab:robustness} shows that under two independent random permutations of the Hard Math Stream, \evopool{} not only maintains its advantage over SA but actually improves slightly, reaching 0.655 and 0.662 compared to 0.639 under the default order. This rules out a favorable curriculum as the explanation: the gains come from the evolution mechanism, not task ordering. We further show in Appendix~\ref{app:more-exp} that varying the initial pool size from $N{=}3$ to $N{=}20$ changes overall accuracy by only 0.011, as lifecycle operators grow or prune the pool to a similar effective size regardless of initialization.

\subsection{Analysis: How the Pool Evolves}
\label{sec:analysis}

\begin{figure}[t]
\centering
\includegraphics[width=0.81\linewidth]{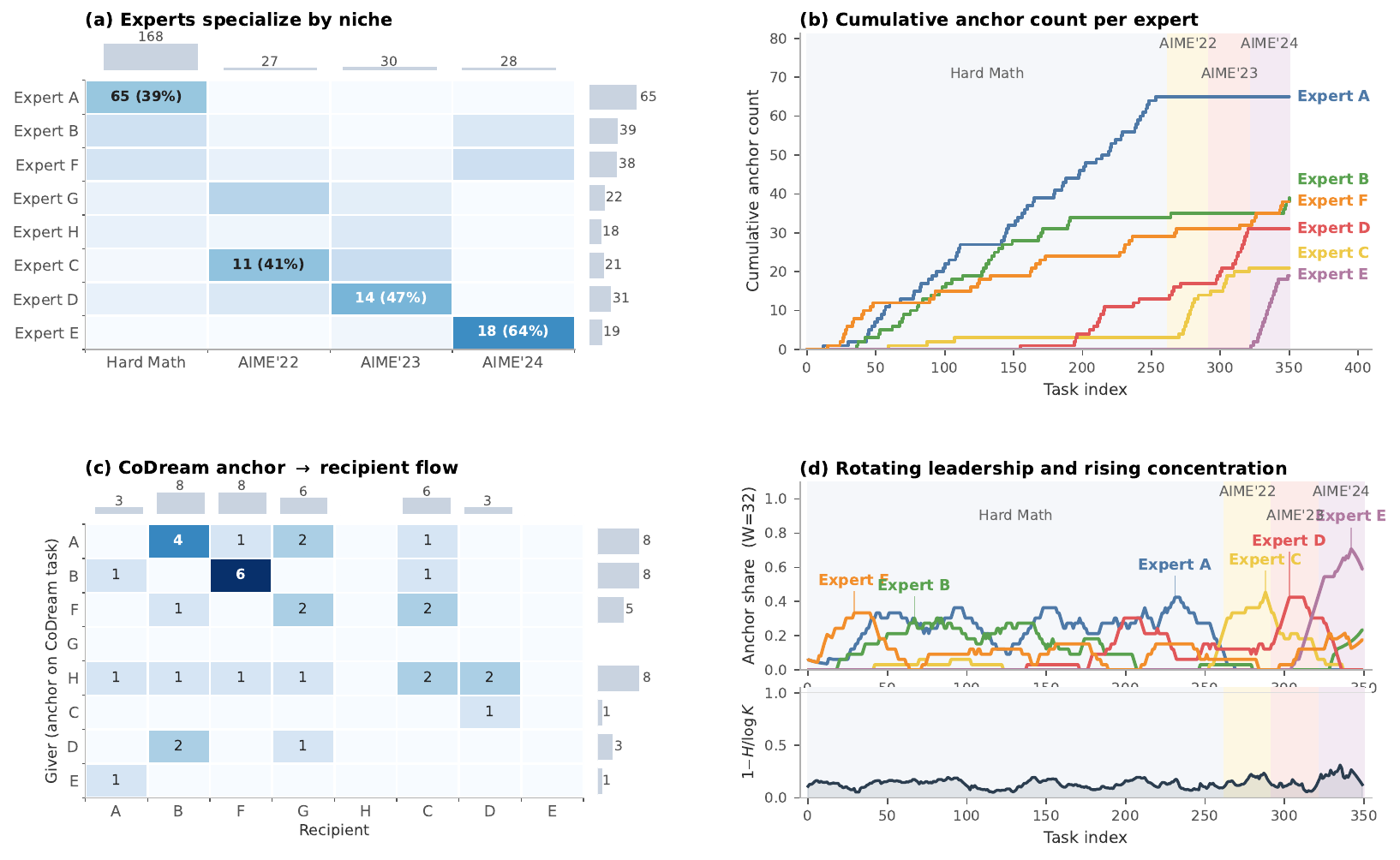}
\caption{Four signals of pool co-evolution on the Hard Math Stream with Qwen3-8B, 382 tasks, seed 42. \textbf{(a)} Expert $\times$ niche anchor counts, column-normalized. \textbf{(b)} Cumulative anchor count per expert across the stream. \textbf{(c)} \codream{} anchor~$\to$~recipient flow matrix. \textbf{(d)} Rolling-window $W{=}32$ anchor share per expert (top) and pool-level specialization index $1{-}H(A^{\text{anchor}}\!\mid\! t)/\log N$ (bottom). For readability we relabel the top eight agents as Expert~A through Expert~H.}
\vspace{-10pt}
\label{fig:evolution-signals}
\end{figure}

Figure~\ref{fig:evolution-signals} reports four signals extracted directly from the run log. Together they show that the pool co-evolves rather than converging to a static assignment, producing phenomena that no single-agent learner can exhibit.

\textbf{Different niches acquire different specialists. } Each niche column converges on a single dominant expert, and the dominant expert differs across AIME years. This niche separation cannot come from the benchmark itself, since all AIME years are math competitions. It falls out of niche-indexed competence $q_i(z)$ updating on task subtype tags.

\textbf{Specialists emerge only when their niche arrives. } math\_hard specialists accumulate from the start, whereas the AIME'23 specialist activates at the AIME'22--'23 boundary and the AIME'24 specialist has zero anchor count until AIME'24 tasks arrive. Specialization is not pre-assigned but surfaces on demand as the competence landscape shifts.

\textbf{Knowledge flows in structured channels, not uniformly. }  \codream{} insights concentrate on a few specific giver~$\to$~recipient cells rather than spreading uniformly. The top givers are the same experts that dominate anchor assignments, and the heavy recipient columns belong to experts that are strong on a different niche, so every expert occupies both roles across the stream.

\textbf{Leadership rotates and concentration rises with task difficulty. } Leaders rotate over the Hard Math phase, and a different expert takes each AIME year. The specialization index rises from $\sim$0.1 on Hard Math toward $\sim$0.3 on AIME'24: the pool concentrates on a single anchor exactly where tasks are hardest.

Taken together, the gains in Tables~\ref{tab:math}--\ref{tab:aflow} do not come from a fixed assignment of experts to niches, but from a continuously updating pool state that routes each task to agents whose competence fits the current niche.

\section{Conclusion}
\label{sec:conclusion}

We have argued that multi-agent test-time evolution is fundamentally different from single-agent evolution replicated $N$ times. Beyond individual context and memory, a multi-agent system evolves who collaborates, how they collaborate, and how knowledge flows across the population. These team and population components have no single-agent counterpart and give rise to emergent phenomena that no individual learner can express. \evopool{} instantiates all three evolution levels over a coevolving agent pool without gradient updates, with \codream{} as its core mechanism for verified asymmetric knowledge transfer. Across three heterogeneous task streams and two model families, \evopool{} consistently outperforms all baselines. The most striking is what emerges without being engineered: N identical agents spontaneously differentiate into several stable niche specialists, leadership rotates across domains, and knowledge flows through structured channels rather than uniformly. This pattern is reproducible across random seeds even as the identity of each specialist changes, confirming emergent specialization as a structural consequence of multi-agent evolution. 


\newpage

\bibliographystyle{plainnat}
\bibliography{references}

\newpage

\appendix

\section*{Appendix}
\section{Limitations and Future Work}
\label{app:limitations}

\textbf{Limitations.} We validate on two model families. Evaluating additional architectures would strengthen generalizability, though we expect the mechanism to transfer since it operates entirely through prompts with no architecture-specific components. The inference cost is roughly $3.6\times$ that of a single agent, which may be prohibitive in latency-sensitive settings, although \evopool{} is more accurate than SC with $k{=}5$ at 72\% of SC's token budget. The lifecycle operators rely on fixed thresholds that transfer across all streams without tuning, but learning them through meta-optimization would be preferable.

\textbf{Future work.} Stronger backbones and longer streams beyond 1000 tasks would enable studies of scaling limits, long-horizon specialization stability, and insight obsolescence. Formalizing role-conditioned credit attribution beyond the current shared team reward is another direction enabled by the three-level decomposition.

\section{More Experiments}
\label{app:more-exp}
\begin{table}[h]
\centering
\caption{Multi-seed specialization metrics on the Hard Math Stream. Across three independent runs with different random seeds, the mean specialization index, peak concentration, and pool expansion (unique anchors from $N{=}20$ initial) are reproducible, while the \emph{identity} of the specialist for each niche changes with the seed. This separates path-dependent identity from seed-invariant pattern.}
\label{tab:multiseed}
\small
\setlength{\tabcolsep}{4pt}
\begin{tabular}{lcccc}
\toprule
Run (seed) & Mean spec. index & Max spec. index & Unique anchors & math\_hard top-1 (distinct?) \\
\midrule
Default order                 & 0.131 & 0.313 & 33 & specialist $\alpha$ (26\%) \\
Shuffle, seed 42              & 0.114 & 0.212 & 42 & specialist $\beta$  (14\%) \\
Shuffle, seed 123             & 0.123 & 0.291 & 40 & specialist $\gamma$ (26\%) \\
\midrule
Mean $\pm$ spread             & $0.123 \pm 0.008$ & $0.272 \pm 0.050$ & $38 \pm 5$ & all three distinct \\
\bottomrule
\end{tabular}
\end{table}

\textbf{Pattern is seed-invariant; identity is seed-dependent.} Table~\ref{tab:multiseed} reports specialization metrics across three independent runs. The specialization index and pool expansion are reproducible across seeds: mean $0.123 \pm 0.008$, unique anchors $38 \pm 5$ from an initial $N{=}20$. However, the specific agent that becomes each niche's specialist is disjoint across the three seeds. The pattern that each niche produces a dominant specialist is a consequence of niche-conditioned selection acting on a shared pool, while the identity of that specialist reflects symmetry breaking at cold start. This separation of seed-invariant pattern from path-dependent identity is a structural signature of multi-agent evolution that single-agent learners cannot produce.

\begin{table}[h]
\centering
\caption{Pool size sensitivity on the Hard Math Stream. Both runs use Qwen3-8B with thinking mode, team size $k{=}3$, seed 42, and identical hyperparameters except the initial pool size $N$.}
\label{tab:poolsize}
\small
\setlength{\tabcolsep}{4.5pt}
\begin{tabular}{lcccccc}
\toprule
Config & math\_hard & AIME'22 & AIME'23 & AIME'24 & AIME'25 & Overall \\
\midrule
$N{=}3$  & 0.740 & 0.433 & 0.333 & 0.433 & 0.333 & 0.628 \\
$N{=}20$ & 0.763 & 0.400 & 0.333 & 0.433 & 0.300 & 0.639 \\
\midrule
$\Delta$ & $-$0.023 & $+$0.033 & 0.000 & 0.000 & $+$0.033 & $-$0.011 \\
\bottomrule
\end{tabular}
\end{table}

\textbf{Pool size has minimal impact on final accuracy.} Table~\ref{tab:poolsize} compares $N{=}3$ and $N{=}20$ on the Hard Math Stream. The overall gap is only 0.011 absolute, concentrated on math\_hard. On all four AIME years, $N{=}3$ matches or slightly exceeds $N{=}20$. The two configurations also converge in pool dynamics: $N{=}3$ grows from 3 to 8 active agents via genesis during the AIME phase, while $N{=}20$ retains only 9 routinely selected agents by the end of the stream, so the effective pool sizes are comparable at convergence. Genesis fires a similar number of times under both configurations, 5 for $N{=}3$ and 4 for $N{=}20$, confirming that lifecycle operators adapt to the current pool state rather than depending on the initial size. This robustness suggests that the evolution mechanism, not the starting roster, is what drives performance.

\section{Related Work Positioning Table}
\label{app:positioning}

Table~\ref{tab:related-app} provides a structured comparison of \evopool{} against representative prior methods along five design axes: whether the method is training-free, whether it maintains a pool of agents, whether knowledge transfers across agents, whether that transfer is asymmetric, and whether evolution is continuous over a task stream. \evopool{} is the only method that satisfies all five criteria simultaneously.

\begin{table}[h]
\centering
\caption{Positioning of \evopool{} against representative prior methods. \checkmark\ = fully satisfied; \xmark\ = not satisfied; $\circ$ = partially satisfied; \textemdash\ = not applicable.}
\label{tab:related-app}
\small
\resizebox{\linewidth}{!}{%
\begin{tabular}{lccccc}
\toprule
\textbf{Method} & \textbf{Training-free} & \textbf{Pool-level} & \textbf{Cross-agent} & \textbf{Asymmetric} & \textbf{Continuous} \\
\midrule
DyLAN, AutoGen, MetaGPT & \checkmark & \checkmark & \textemdash & \textemdash & \xmark \\
AFlow, ScoreFlow          & \checkmark$^\dagger$ & \xmark & \textemdash & \textemdash & \xmark \\
Reflexion, MemGPT~\citep{memgpt2023}, EvoAgent & \checkmark & \xmark & \xmark & \textemdash & \checkmark \\
AgentNet                  & \checkmark & \checkmark & \xmark & \textemdash & \checkmark \\
EvoMem (pool Reflexion)   & \checkmark & \checkmark & \xmark & \textemdash & \checkmark \\
MemCollab                 & \checkmark & \checkmark & \checkmark & \xmark & \checkmark \\
MAS\textsuperscript{2}    & \xmark          & \checkmark & \checkmark & $\circ$ & \checkmark \\
\midrule
\textbf{\evopool{} (full)} & \textbf{\checkmark} & \textbf{\checkmark} & \textbf{\checkmark} & \textbf{\checkmark} & \textbf{\checkmark} \\
\bottomrule
\end{tabular}%
}
\vspace{0.3em}

{\footnotesize $\dagger$AFlow's MCTS requires hundreds of offline LLM calls per domain; the resulting workflow is frozen at inference time.}
\end{table}

\section{Experience Archive Design Justification}
\label{app:memory-tiers}

As described in Section~\ref{sec:individual}, each agent maintains two stores that separate reasoning insights by scope.

\textbf{Subtask-level lessons} are indexed by niche label and capture domain-scoped strategies, such as a proof technique for combinatorics or a debugging pattern for recursive algorithms. These lessons are retrieved by cosine similarity over task embeddings when the agent encounters a task in the same niche, providing targeted in-context guidance. Near-duplicate entries are merged via LLM-based deduplication to control redundancy.

\textbf{Cross-domain meta-insights} form a single pool not tied to any niche, capturing higher-order self-corrections such as ``decompose the problem into sub-steps independently.'' Without a dedicated cross-domain store, an agent that learns careful decomposition from math cannot transfer this principle to code or QA without re-discovering it.

Both stores grow with the agent's full history, with no fixed capacity limit. At solve time, entries from both stores are retrieved by cosine similarity and prepended to the prompt as in-context guidance.

Separating niche-specific from cross-domain insights serves two purposes. First, it prevents tactical noise from polluting general metacognition. Second, it gives \codream{} the granularity needed to route each insight to the right audience: niche-local strategies are sent only to deficit agents on that niche, while cross-domain insights can propagate more broadly.
\section{Implementation Details}
\label{app:impl}

\subsection{Operational Details}
\label{app:operational}

We provide concrete definitions for the quantities referenced in \S\ref{sec:method}.

\textbf{Style overlap $\omega_{ij}$.} Each agent maintains a dictionary mapping subdomain tags to its running competence on that subdomain. The style overlap $\omega_{ij}$ is the cosine similarity of these two competence vectors, aligned over the union of both agents' subdomain keys. When a subdomain tag appears in one agent's dictionary but not the other, the missing entry is treated as zero competence. High overlap indicates that two agents have developed similar skill profiles across the same set of subdomains, meaning they would contribute redundant perspectives to a team. Low overlap can arise either because the agents specialize in different subdomains or because one agent has been exposed to subdomains that the other has not encountered.

\textbf{Pair synergy $\sigma_{ij}(z)$.} The pair synergy on niche $z$ is the mean team reward on past niche-$z$ tasks in which agents $i$ and $j$ both participated. It is initialized to $0$ and remains at $0$ until the pair has co-participated in at least five niche-$z$ tasks, avoiding noisy estimates from small samples. Synergy captures whether two agents complement each other on a specific niche: a pair that consistently achieves higher team rewards than either agent's solo competence would predict has high synergy.

\textbf{Lifecycle operators.} All four operators are evaluated every $\tau = 10$ tasks.

\emph{Fork} targets agents in the top 10\% by rolling-average reward. The operator clones the selected agent and mutates the clone's persona via a one-shot LLM call that instructs the backbone to emphasize the parent's dominant subdomain while preserving the parent's general role description. The clone inherits a copy of the parent's full memory store but receives a distinct agent ID, so subsequent competence updates diverge. Fork serves as controlled exploration in persona space: it amplifies successful strategies while introducing variation that may discover adjacent niches.

\emph{Merge} fires when a pair's profile cosine similarity exceeds $0.95$ and both agents have accumulated at least 10 tasks. The two agents are consolidated into a single agent that inherits both memory stores, with near-duplicate entries deduplicated via the same LLM-based deduplication used during normal insight injection.

\emph{Prune} retires agents that have scored below $0.8\times$ the pool mean for 10 or more consecutive tasks. Pruned agents are removed from the pool entirely and their memory stores are discarded.

\emph{Genesis} is triggered when the pool size drops below 15 agents or when no existing agent has niche affinity greater than $0.4$ on a newly encountered task type. New agents are seeded with domain-specific personas generated by an LLM call that describes the uncovered niche, but with empty memory stores, so all subsequent knowledge must be earned through task experience.

\subsection{Inference Configuration}
\label{app:inference}

\textbf{Serving.} All experiments use Qwen3-8B served locally via vLLM~\cite{vllm} with two instances under round-robin load balancing. Key parameters: tensor parallel size 1 GPU per instance, max model length 32,768 tokens, GPU memory utilization 0.90, max batch size 32, dtype bfloat16.

\textbf{Generation mode.} All three streams use thinking mode.

\textbf{Token budgets.} Task solving uses 4,096 output tokens per agent, increased to 8,192 for Hard Math with thinking. Each \codream{} phase uses 2,048 tokens per agent. Profile injection prepends retrieved insights to the system prompt.

\subsection{Hyperparameters}
\label{app:hyperparams}

Table~\ref{tab:hyperparams} lists all hyperparameters. A single configuration is used across all three streams and both model families with no per-benchmark tuning.

\begin{table}[h]
\centering
\caption{\evopool{} hyperparameters. A single configuration is used across all streams and backbones.}
\label{tab:hyperparams}
\small
\resizebox{\linewidth}{!}{%
\begin{tabular}{lll}
\toprule
Hyperparameter & Value & Notes \\
\midrule
Pool size $N$ & 20 & Fixed across all streams \\
Team size $k$ & 3 & Greedy one-at-a-time selection \\
EWMA decay $\alpha$ & 0.3 & For $q_i(z)$ competence update \\
$q_i(z)$ initialization & 0.5 & Prior to first niche encounter \\
Complement weights $(\lambda_q, \lambda_\sigma, \lambda_\omega)$ & $(1.0, 0.3, 0.5)$ & Competence / synergy / style-overlap penalty \\
Scout weights $(\lambda_u, \lambda_d)$ & $(0.3, 0.5)$ & Under-exposure / diversity penalty \\
Lifecycle interval $\tau$ & 10 tasks & Fork / merge / prune / genesis check \\
Fork threshold & Top 10\% by rolling average & Specializes high performers \\
Merge threshold & Profile cosine sim $>0.95$, $\ge$10 tasks each & Collapses redundant agents \\
Prune threshold & Below $0.8\times$ pool mean for $\ge$10 consecutive tasks & Retires persistent underperformers \\
Genesis trigger & Max niche affinity $<$0.4 & Seeds domain-specific new agents \\
\codream{} trigger $\theta$ & 0.6 & Team reward threshold \\
Insight dedup cosine & 0.85 & Prevents near-duplicate insights \\
Deficit gate & Below-median recent performance & For asymmetric routing \\
\bottomrule
\end{tabular}%
}
\end{table}

\subsection{Evaluation Protocol}
\label{app:eval-protocol}

All streams use a fixed task order across methods to ensure comparable learning trajectories.

All agents are initialized with a generic helpful-assistant persona; domain-specific knowledge emerges entirely from task experience.

\section{\codream{} Isolation Experiment}
\label{app:isolation}

The main-text ablation in Table~\ref{tab:ablation} removes \codream{} from the full system while keeping all other components. Here we run a more controlled isolation on a smaller scale to sharpen the conclusion. We select a 30-task math subsequence from AFlow-Stream and compare three configurations that differ in exactly one dimension: SA uses a single agent with no pool, \evopool{} w/o \codream{} maintains the full 20-agent pool with individual experience accumulation, team composition, and lifecycle operators but disables cross-agent knowledge transfer, and \evopool{} (full) enables \codream{} on top of the same pool infrastructure. The goal is to test whether pool infrastructure alone already improves over a single agent, or whether the improvement requires cross-agent transfer.

\begin{table}[h]
\centering
\caption{Controlled \codream{} isolation on a 30-task AFlow math subsequence.}
\label{tab:isolation}
\small
\begin{tabular}{lc}
\toprule
Configuration & Accuracy \\
\midrule
SA (no pool)                              & 0.633 \\
\evopool{} w/o \codream{}                 & 0.633 \\
\textbf{\evopool{} (full)}                & \textbf{0.700} \\
\bottomrule
\end{tabular}
\end{table}

The key finding is that \evopool{} without \codream{} matches SA exactly at 0.633. Maintaining 20 agents with individual experience, team composition, and lifecycle management produces zero gain over a single agent when cross-agent knowledge sharing is absent. This result is expected on a short, single-domain subsequence: without \codream{}, each agent accumulates experience independently, and the team composition operator can select competent agents but cannot transfer knowledge from strong agents to weak ones. The pool infrastructure provides the scaffolding for knowledge flow, but it is \codream{} that activates the flow.

Adding \codream{} yields 0.700, a $+$10.5\% relative improvement, confirming that asymmetric transfer is the mechanism responsible for the multi-agent advantage on this subset. On the full 600-task AFlow-Stream, the gap is even larger: removing \codream{} causes a $-$0.108 drop in overall accuracy (Table~\ref{tab:ablation}), with the effect concentrated on dependent-reasoning domains such as HotpotQA and DROP where cross-agent coordination knowledge is most valuable.

\section{Hard Code Stream Per-Benchmark Breakdown}
\label{app:code}
\label{app:code-humaneval}

Table~\ref{tab:code} in the main text omits the HumanEval column because HumanEval saturates at 1.000 for every method in our harness. Overall is the micro-average over all 586 tasks including HumanEval. Table~\ref{tab:code-app} provides the full per-benchmark breakdown.

\begin{table}[h]
\centering
\caption{Hard Code Stream per-benchmark accuracy. HumanEval saturates at ceiling for all methods. Overall is the micro-average over all 586 tasks, matching Table~\ref{tab:code}.}
\label{tab:code-app}
\small
\begin{tabular}{lcccc}
\toprule
Method & MBPP+ & HumanEval & CodeContests & Overall \\
\midrule
SA                        & 0.842 & 1.000 & 0.068 & 0.667 \\
SC ($k{=}5$)              & 0.849 & 1.000 & 0.198 & 0.708 \\
DyLAN                     & 0.825 & 1.000 & 0.189 & 0.695 \\
AgentNet                  & \textbf{0.887} & 1.000 & 0.102 & 0.698 \\
EvoMem                    & 0.885 & 1.000 & 0.027 & 0.672 \\
MemCollab                 & 0.870 & 1.000 & 0.084 & 0.682 \\
\midrule
\rowcolor{cyan!5} \textbf{\evopool{} (full)}  & 0.861 & 1.000 & \textbf{0.352} & \textbf{0.757} \\
\bottomrule
\end{tabular}
\end{table}

MBPP+ clusters near 0.85 for all multi-agent methods, leaving CodeContests as the discriminating subset. We inspected HumanEval task-level outputs and confirmed that all methods solve every problem correctly; the remaining minor diversity across runs is within grading tolerance.

On CodeContests, \evopool{} achieves 0.352, a $1.8\times$ improvement over SC $k$=5 and $3.5\times$ over AgentNet. The mechanism is experience-guided debugging: agents whose experience archives contain prior failure patterns and repair strategies for similar problem classes attempt more targeted corrections on subsequent CodeContests problems. This is a direct consequence of cross-difficulty transfer within the stream, as debugging patterns first learned on easier MBPP+ problems accumulate in agent profiles and propagate to deficit agents via \codream{} before the harder CodeContests problems arrive.

\section{Order and Execution Robustness: Setup}
\label{app:robust-config}

This section provides setup details for the robustness experiments reported in Table~\ref{tab:robustness}. All runs use the same Qwen3-8B backbone, the same pool and team sizes, and the same code version as the main Hard Math Stream result in Table~\ref{tab:math}.

\textbf{Shuffle conditions.} The default task order presents domains in sequential blocks as described in \S\ref{sec:setup}. The two shuffle conditions reorder all 382 tasks across domains using the given random seed, producing a different task ordering while preserving the same task set.

\textbf{Uniform execution.} This condition disables \textsc{LeadLearn}'s dynamic structure selection and forces voting for every task, using the same self-consistency implementation as the SC baseline for each team member. All other components remain intact: individual experience accumulation, \codream{}, and lifecycle operators.

\textbf{SA reference.} The SA score in the table is the default fixed-order result. Since SA does not accumulate experience across tasks, its performance under any task permutation is statistically indistinguishable from the fixed-order score. We verified this on a single shuffled run with seed 42, which produced equivalent results.

\section{Per-Subset Regime Analysis}
\label{app:regime}

\evopool{}'s per-subset gains vary by the per-agent success rate on that subset, with a regime structure that matches the underlying mechanism.

\textbf{Very high accuracy ($\ge$80\%).}
When the backbone already solves most tasks, such as GSM8K, MBPP, or MATH Level~3 with strong backbones like GPT-4.1-mini on math\_hard at 0.824, there is little room for improvement and gains are modest, ranging from $+0.01$ to $+0.07$.
In this regime, the dominant contribution comes from team diversity and the leader's dynamic structure selection rather than from cross-agent distillation.

\textbf{Mid accuracy (40\%--70\%).}
MATH Level~4/5 with Qwen3-8B, where math\_hard base accuracy is 0.302, is the design sweet spot for \codream{}: enough verified solutions for reliable crystallization, yet wide gaps between struggling and successful agents.
Gains here range from $+0.07$ on Qwen3-8B math\_hard to $+0.20$+ on GPT-4.1-mini AIME'22 where base accuracy is 0.40, and the cross-agent distillation component contributes a meaningful share.

\textbf{Low accuracy (20\%--40\%).}
AIME-level tasks under Qwen3-8B yield per-agent success rates near 15--20\%, and under GPT-4.1-mini near 30--40\%.
Verified solutions are rare but not absent.
For Qwen3-8B, \codream{}'s direct contribution on AIME is small and statistically noisy at this backbone capability. The full \evopool{} system still lifts AIME via team selection and lifecycle.
For GPT-4.1-mini, the same AIME subsets land in a higher base-accuracy regime of 30--40\%, and we observe the largest per-subset gains of the entire paper, $+0.20$ to $+0.27$.
This is consistent with the mechanism: with more frequent verified solutions, cross-agent distillation has more material to crystallize and route.

\textbf{Very low accuracy ($<$15\%).}
No effect is expected: at 10\% per-agent accuracy with team of 3, the probability that at least one agent succeeds is $1 - (0.90)^3 = 0.271$, and the probability of a verified solution on two independent attempts drops quickly.
In practice we observe \codream{} correctly abstains when no agent solves a task.

\textbf{Self-Consistency collapse derivation.}
With five independent agents at 20\% accuracy, the probability of majority-correct:
\[
  P(\text{majority correct}) = \sum_{j=3}^{5}\binom{5}{j}(0.20)^j(0.80)^{5-j}
  \approx 0.058 = 5.8\%.
\]
This is \emph{lower} than 20\% single-agent: majority voting actively overrides rare correct answers.
The empirical SC result of 0.067 on AIME matches this prediction.
\evopool{} avoids this failure mode because its team leader selects non-voting structures such as debate, generator-critic, or decompose when the base rate is low and rare successes exist.

\section{Case Study: How \evopool{} Learns Competition Mathematics}
\label{app:case-study}

We trace specific events from the Hard Math Stream rerun used in Table~\ref{tab:math}, conducted with Qwen3-8B on seed 42 over 382 tasks.
All events below are parsed directly from the per-task run log. Agent IDs are real, truncated to 8 hex characters.

\textbf{Early expert identification, tasks 10--30 on math\_hard.}
The first lifecycle events fire at task~10, after the rolling performance window has stabilized.
All three events are simultaneous \emph{specialize}/fork events on agents \texttt{6f3dcc14}, \texttt{bb411e98}, and \texttt{119b9e09}, each with sustained mean reward of 0.80--1.00 on math\_competition\_hard.
These same three agents are forked again at tasks~20, 30, 50, and most other lifecycle checkpoints in the math\_hard phase.
By the end of the math\_hard phase at task~261, these three account for all 43 specialize events in the entire stream.

\textbf{Insight crystallization on math\_hard.}
The first \codream{} trigger occurs at task~11 with team score 0/3.
Three insights are crystallized simultaneously by the same three agents \texttt{6f3dcc14}, \texttt{bb411e98}, \texttt{119b9e09}.
A representative insight:
\begin{quote}
\emph{``When counting integers in a range divisible by multiple numbers, use the inclusion-exclusion principle with LCM adjustments: (1) compute LCM of all divisors; (2) use inclusion-exclusion to count numbers divisible by subsets of divisors; (3) alternate signs.''}
\end{quote}
Over the math\_hard phase, these same three agents produce 72 of 93 verified insights in the run, or 77\% of the total.
The insights are concrete competition-math techniques: generating functions for recursive growth, Möbius inversion for overlapping-set counts, Chinese Remainder Theorem for systems of congruences.

\textbf{Quality degradation on AIME.}
\codream{} fires 7 times in the AIME phase, 4 on AIME'22 and 3 on AIME'23, with zero triggers on AIME'24 or AIME'25. The character of the insights changes.
At task 267, \texttt{aime\_2022\_5} with team score 0, the three insights crystallized are not math strategies but meta-advice about extracting numerical values from text:
\begin{quote}
\emph{``When a problem involves extracting and reconciling numerical data from a passage with multiple steps or implicit relationships, create a structured checklist of required values \ldots''}
\end{quote}
When no team member solves the underlying math problem, the crystallize step has no successful trajectory to distill from, and the agents' reflections produce generic reading-comprehension advice rather than targeted mathematical techniques.
This is consistent with \codream{}'s regime condition in \S\ref{app:regime}: on very hard problems where per-agent accuracy is too low for any team member to succeed, the mechanism cannot extract useful material.
The verification gate still passes these candidates because the re-attempt with meta-advice applied happens to score marginally higher than the original failure, but their contribution to future AIME performance is marginal.

\textbf{Late-stream lifecycle: from forking to genesis.}
Specialize events stop entirely at the math\_hard/AIME boundary, task 262.
In the AIME phase, the system instead fires 5 \emph{genesis} events seeding new agents in response to ``coverage gap for task type \texttt{aime\_problem}, max affinity = 0.20'' and 1 \emph{prune} removing an agent with 6 consecutive underperforming tasks.
\textbf{Summary: emergent structure from identical initialization.}
All 20 agents start from the same backbone with empty insight stores.
Over 382 tasks, \codream{} fires 34 times and crystallizes 93 verified insights.
These insights are not evenly distributed: seven agents contribute all 93, and the top three, \texttt{bb411e98}, \texttt{119b9e09}, and \texttt{6f3dcc14}, contribute 72 of 93, or 77\% of the total.
Independently, lifecycle specialization events concentrate on the same three agents: these three account for all 43 specialize events, with \texttt{bb411e98} at 16, \texttt{6f3dcc14} at 14, and \texttt{119b9e09} at 13.
The top insight contributors and the top forked agents overlap completely. An expert core differentiates from the pool purely through environment feedback.

The system's lifecycle behavior also splits cleanly by regime: all 43 specialize events occur during the math\_hard phase where per-agent base accuracy is sufficient to identify clear top performers and fork them for controlled exploration.
During the AIME phase, specialize events cease and lifecycle shifts to a different mix: 5 genesis events for coverage gaps and 1 prune for consecutive underperformance.
This mirrors the regime analysis in \S\ref{app:regime}: at mid base accuracy the system can identify and amplify specialists, while at very low base accuracy it instead seeds new agents and retires unproductive ones.

\section{Lifecycle Operator Analysis}
\label{app:lifecycle}

\begin{table}[h]
\centering
\caption{Lifecycle contribution by stream phase on AFlow-Stream.}
\label{tab:lifecycle-phase}
\begin{tabular}{lccc}
\toprule
Configuration & Early (1--200) & Mid (201--400) & Late (401--600) \\
\midrule
\evopool{} (full)    & 0.868 & 0.876 & 0.879 \\
$-$Lifecycle         & 0.867 & 0.869 & 0.871 \\
$\Delta$             & $+$0.001 & $+$0.007 & $+$0.008 \\
\bottomrule
\end{tabular}
\end{table}

Lifecycle contribution is negligible early and grows modestly in mid-to-late phases.
Fork/merge/retire maintain diversity and prune stagnation over long streams, not accelerate early learning.



\section{\codream{} Insight Examples}
\label{app:insights}

Representative insights crystallized during actual experimental runs, lightly edited for brevity. Each insight was generated by a specific agent during a post-task reflection session, verified by re-attempt, and routed to the appropriate experience archive.

\subsection{Math Insights, Hard Math Stream, Qwen3-8B}

\textbf{Example M1: modular arithmetic, from task math\_hard\_10.}
\begin{quote}
\itshape
``Modular arithmetic constraints must be integrated into the sequence's structural definition rather than treated as external constraints. This integration allows for a more accurate modeling of sequences where the modulus influences the sequence's recursive or periodic behavior.''
\end{quote}

\textbf{Example M2: trapezoid geometry, from task math\_hard\_13.}
\begin{quote}
\itshape
``The correct application of the trapezoid area formula hinges not just on identifying parallel sides, but also on accurately measuring the perpendicular height. A structured geometric analysis, starting with side identification, followed by precise height measurement, and finally applying the formula, prevents formula misapplication.''
\end{quote}

\textbf{Example M3: constraint-graph reformulation, from AIME 2022.}
\begin{quote}
\itshape
``For constraint-satisfaction problems, model as a graph where nodes represent constraints and edges represent interactions; this allows more efficient traversal and resolution of complex dependencies than direct decomposition.''
\end{quote}

\subsection{Code Insights, Hard Code Stream, Qwen3-8B}

\textbf{Example C1: memoization with state compression, from MBPP task~54.}
\begin{quote}
\itshape
``When a combinatorial problem has high symmetry or complex dependencies, use memoization with state compression: represent the state as a tuple of essential parameters, and cache results to avoid redundant computation.''
\end{quote}

\textbf{Example C2: bounded arithmetic with saturation, from CodeContests task~425.}
\begin{quote}
\itshape
``When input validation involves numerical ranges and potential overflow, use bounded arithmetic with explicit saturation: clamp intermediate values to the valid range [min\_val, max\_val] using \texttt{min(max(value, min\_val), max\_val)} before further computation.''
\end{quote}

\textbf{Example C3: symbolic + numerical cross-validation, from MBPP task~17.}
\begin{quote}
\itshape
``When symbolic computation verifies mathematical logic with potential edge cases, cross-validate with numerical evaluation at specific test points: (1) define test points covering edge cases and typical scenarios; (2) evaluate the symbolic expression numerically; (3) compare symbolic and numerical results to catch simplification errors.''
\end{quote}

These examples illustrate the style of crystallized insights: actionable, cross-task patterns rather than problem-specific hints.
In the current implementation, most insights are classified as cross-domain by the crystallization step. Niche-specific routing is exercised when the insight-classification prompt assigns lower transferability, and the asymmetric sharing decision between selective and broadcast routing is additionally gated by the cosine-similarity deficit check at injection time.


\newpage

\end{document}